\documentclass[11pt,a4paper]{article}
\usepackage[hyperref]{emnlp2018}
\usepackage{times}
\usepackage{latexsym}
\usepackage{url}
\usepackage{amsmath,amssymb,amsfonts}
\usepackage{threeparttable}
\usepackage{graphicx}

\aclfinalcopy % Uncomment this line for the final submission
\title{Neural Related Work Summarization with a Joint Context-driven Attention Mechanism}

\author{Yongzhen Wang$^1$\thanks{\;\;Corresponding author},\quad Xiaozhong Liu$^{2,3}$,\quad Zheng Gao$^2$\\
$^1$School of Maritime Economics and Management, Dalian Maritime University, Dalian, China\\
$^2$School of Informatics, Computing and Engineering, Indiana University Bloomington,\\
Bloomington, IN, USA\\
$^3$Alibaba Group, Hangzhou, China\\
{\tt \small $^*$kuadmu@163.com} \quad
{\tt \small liu237@indiana.edu} \quad
{\tt \small gao27@umail.iu.edu}
}
\date{}

\begin{document}
\maketitle
\begin{abstract}
    Conventional solutions to automatic related work summarization rely heavily on human-engineered features. In this paper, we develop a neural data-driven summarizer by leveraging the seq2seq paradigm, in which a joint context-driven attention mechanism is proposed to measure the contextual relevance within full texts and a heterogeneous bibliography graph simultaneously. Our motivation is to maintain the topic coherency between a related work section and its target document, where both the textual and graphic contexts play a big role in characterizing the relationship among scientific publications accurately. Experimental results on a large dataset show that our approach achieves a considerable improvement over a typical seq2seq summarizer and five classical summarization baselines.
\end{abstract}

\section{Introduction}

In scientific fields, scholars need to contextualize their contribution to help readers acquire an understanding of their research papers. For this purpose, the related work section of an article serves as a pivot to connect prior domain knowledge, in which the innovation and superiority of current work are displayed by a comparison with previous studies. While citation prediction can assist in drafting a reference collection \citep{Nallapati2008Joint}, consuming all these papers is still a laborious job, where authors must read every source document carefully and locate the most relevant content cautiously.

As a solution in saving authors' efforts, automatic related work summarization is essentially a topic-biased multi-document problem \citep{Cong2010Towards}, which relies heavily on human-engineered features to retrieve snippets from the references. Most recently, neural networks enable a data-driven architecture {\it sequence-to-sequence} (seq2seq) for natural language generation \citep{Bahdanau2014Neural,Bahdanau2016End}, where an encoder reads a sequence of words/sentences into a context vector, from which a decoder yields a sequence of specific outputs. Nonetheless, compared to scenarios like machine translation with an end-to-end nature, aligning a related work section to its source documents is far more challenging.

To address the summarization alignment, former studies try to apply an attention mechanism to measure the saliency/novelty of each candidate word/sentence \citep{Tan2017Abstractive}, with the aim of locating the most representative content to retain primary coverage. However, toward summarizing a related work section, authors should be more creative when organizing text streams from the reference collection, where the selected content ought to highlight the topic bias of current work, rather than retell each reference in a compressed but balanced fashion. This motivates us to introduce the contextual relevance and characterize the relationship among scientific publications accurately.

Generally speaking, for a pair of documents, a larger lexical overlap often implies a higher similarity in their research backgrounds. Yet such a hypothesis is not always true when sampling content from multiple relevant topics. Take ``DSSM"\footnote{Learning deep structured semantic models for web search using clickthrough data \citep{huang2013learning}} as an example, from viewpoint of the abstract similarity, those references investigating ``Information Retrieval", ``Latent Semantic Model" or ``Clickthrough Data Mining" could be of more importance in correlation and should be greatly sampled for the related work section. But in reality, this article spends a bit larger chunk of texts (about 58\%) to elaborate ``Deep Learning" during the literature review, which is quite difficult for machines to grasp the contextual relevance therein. In addition, other situations like emerging new concepts also suffer from the terminology variation or paraphrasing in varying degrees.

In this study, we utilize a heterogeneous bibliography graph to embody the relationship within a scalable scholarly database. Over the recent past, there is a surge of interest in exploiting diverse relations to analyze bibliometrics, ranging from literature recommendation \citep{Yu2015Random} to topic evolvement \citep{Jensen2016Generation}. In a graphical sense, interconnected papers transfer the credit among each other directly/indirectly through various patterns, such as paper citation, author collaboration, keyword association and releasing on series of venues, which constitutes the graphic context for outlining concerned topics. Unfortunately, a variety of edge types may pollute the information inquiry, where a slice of edges are not so important as the others on sampling content. Meanwhile, most existing solutions in mining heterogeneous graphs depend on the human supervision, e.g., {\it hyperedge} \citep{Bu2010Music} and {\it metapath} \citep{Swami2017metapath2vec}. This is usually not easy to access due to the complexity of graph schemas.

Our contribution is threefold: First, we explore the {\it edge-type usefulness distribution} (EUD) on a heterogeneous bibliography graph, which enables the relationship discovery (between any pair of papers) for sampling the interested information. Second, we develop a novel seq2seq summarizer for the automatic related work summarization, where a joint context-driven attention mechanism is proposed to measure the contextual relevance within both textual and graphic contexts. Third, we conduct experiments on 8,080 papers with native related work sections, and experimental results show that our approach outperforms a typical seq2seq summarizer and five classical summarization baselines significantly.
\section{Related Work}

This study touches on several strands of research within automatic related work summarization and seq2seq summarizer as follows.

The idea of creating a related work section automatically is pioneered by \citet{Cong2010Towards} who design two rule-based strategies to extract sentences for general and detailed topics respectively. Subsequently, \citet{Hu2014Automatic} exploit {\it probabilistic latent semantic indexing} to split candidate texts into different topic-biased parts, then apply several regression models to learn the importance of each sentence. Similarly, \citet{Widyantoro2014Citation} transform the summarization problem into classifying rhetorical categories of sentences, where each sentence is represented as a feature vector containing word frequency, sentence length and etc. Most recently, \citet{Chen2016Summarization} construct a graph of representative keywords, in which a {\it minimum steiner tree} is figured out to guide the summarization as finding the least number of sentences to cover the discriminated nodes. In general, compared to traditional summaries, the automatic related work summarization receives less concerns over the past. However, these existing solutions cannot work without manual intervention, which limits the application scale to an extremely small size (see Table \ref{table:1}).

\begin{table}[t!]
\small
    \begin{center}
        \begin{tabular}{|c|c|}
        \hline Authors & Number of papers \\ \hline
        \citet{Cong2010Towards} & 20 \\
        \citet{Hu2014Automatic} & 1,050 \\
        \citet{Widyantoro2014Citation} & 50 \\
        \citet{Chen2016Summarization} & 3 \\
        \hline
        \end{tabular}
    \end{center}
\caption{\label{table:1} Data scales of previous studies on automatic related work summarization.}
\end{table}

The earliest seq2seq summarizer stems from \citet{Rush2015A} which utilizes a feed-forward network for compressing sentences, and later is expanded by \citet{Chopra2016Abstractive} with a {\it recurrent neural network} (RNN). On this basis, \citet{Nallapati2016Sequence,Nallapati2016Abstractive} and \citet{Chen2016Distraction} both present a set of RNN-based models to address various aspects of abstractive summarization. Typically, \citet{Cheng2016Neural} propose a general seq2seq summarizer, where an encoder learns the representation of documents while a decoder generates each word/sentence using an attention mechanism. With further research, \citet{Nallapati2016SummaRuNNer} extend the sentence compression by trying a hierarchical attention architecture and a limited vocabulary during the decoding phase. Next, \citet{Narayan2017Neural} leverage the side information as an attention cue to locate focus regions for summaries. Recently, inspired by {\it PageRank}, \citet{Tan2017Abstractive} introduce a graph-based attention mechanism to tackle the saliency problem. Nonetheless, these methods all discuss the single-document scenario, which is far from the nature of automatic related work summarization.

In this study, derived from the general seq2seq summarizer of \citet{Cheng2016Neural}, we propose a joint context-driven attention mechanism to measure the contextual relevance within full texts and a heterogeneous bibliography graph simultaneously. To our best knowledge, we make the first attempt to develop a neural data-driven solution for the automatic related work summarization, and the practice of using the joint context as an attention cue is also less explored to date. Besides, this study is launched on a dataset with up to 8,080 papers, which is much greater than previous studies and makes our results more convincing.

Since text summarization via word-by-word generation is not mature at present \citep{Cheng2016Neural,Nallapati2016SummaRuNNer,Tan2017Abstractive}, we adopt the extractive sentential fashion for our summarizer, where a related work section is created by extracting and linking sentences from a reference collection. Meanwhile, this study follows the mode of \citet{Cong2010Towards} who assume that the collection is given as part of the input, and do not consider the citation sentences of each reference.
\section{Methodology}

\subsection{Problem Formulation}

To adapt the seq2seq paradigm, we formulate the automatic related work summarization into a sequential text generation problem as follows.

Given an unedited paper ${\tt t}$ (target document) and its $n$-size reference collection ${\tt R}_{\tt t}=\{ {\tt r}_{1:n}^{\tt t}\}$, we draw up a related work section for ${\tt t}$ by selecting sentences from ${\tt R}_{\tt t}$. To be specific, each reference (source document) will be traversed one time sequentially, and without loss of generality, in the descending order of their significance to ${\tt t}$. Consequently, all sentences to be selected are concatenated into an $m$-length sequence ${\tt S}_{\tt t}=\{{\tt s}_{1:m}^{\tt t}\}$ to feed the summarizer. For each candidate sentence ${\tt s}_j^{\tt t}$, once being visited, a label ${\tt y}_j^{\tt t}\in\{0,1\}$ will be determined synchronously based on whether or not this sentence should be covered into the output. Our objective is to maximize the log-likelihood probability of observed labels ${\tt Y}_{\tt t}=\{{\tt y}_{1:m}^{\tt t}\}$ under ${\tt R}_{\tt t}$, ${\tt S}_{\tt t}$ and summarizer parameters $\theta$, as shown below.

\begin{equation}
    \max\sum_{j=1}^m\log\text{Pr}({\tt y}_j^{\tt t}\mid{\tt R}_{\tt t};{\tt S}_{\tt t};\theta)
\end{equation}

\subsection{Random Walk on Heterogeneous Bibliography Graph}

\begin{figure}[htb]
\centering
    \includegraphics[width=0.4\textwidth]{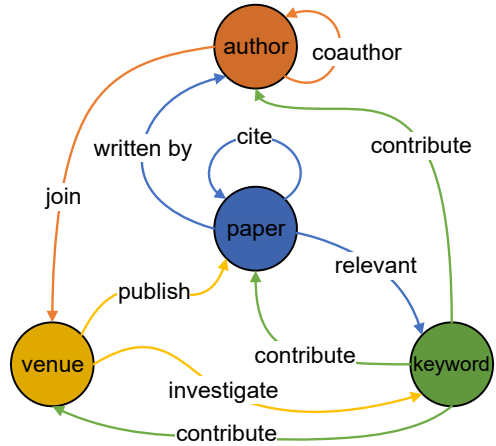}
\caption{Heterogeneous bibliography graph.}
\label{fig:1}
\end{figure}

Prior works have illustrated that one of the most promising channels for information recommendation is the community network \citep{Guo2015Automatic}. In this study, we verify this hypothesis toward the content sampling of scientific summarization, by investigating heterogeneous relations among different kinds of objects such as papers, authors, keywords and venues.

For measuring the relationship among scientific publications, we introduce a directed graph ${\tt G}=({\tt V},{\tt E})$ to contain various bibliographical connections, as shown in Figure \ref{fig:1}, which involves four objects and ten edge types in total. Each edge ${\tt e}_{j,i}\in{\tt E}$ is assigned a value $\frac{\pi({\tt e}_{j,i})}{z}\in[0,1]$ to indicate the transition probability between two nodes ${\tt v}_j,{\tt v}_i\in{\tt V}$, where $\pi({\tt e}_{j,i})\in\mathbb{R}$ returns the unknown edge-type usefulness of ${\tt e}_{j,i}$, and $z\in\mathbb{R}$ is a normalizing weight. For most of edge types, we model the weight as one divided by the number of outgoing links of the same kind. But regarding the ``contribution" category, the weight modeling is accomplished by {\it PageRank with Priors} \citep{White2003Algorithms}. Note that different edge types usually take very uneven importance in one particular task \citep{Yu2015Random}, and it is quite difficult to enable the classical heterogeneous graph mining without expert defined paths for {\it random walk} \citep{Bu2010Music,Swami2017metapath2vec}.

In this study, we propose an unsupervised approach to capture the connectivity diversity, by introducing an optimal EUD for navigating {\it random walkers} on the heterogeneous bibliography graph. Given a target document ${\tt t}$, the optimized usefulness assignment can help those walkers lock a top-$n$ recommendation $\bar{{\tt R}}_{\tt t}$ to best match the reference collection ${\tt R}_{\tt t}$, as shown in Eq. \ref{eq:2}. On this basis, a well-performing algorithm {\it node2vec} \citep{Grover2016node2vec} is adopted to conduct an unsupervised {\it random walk} to vectorize every node $\forall{\tt v}_*\in{\tt V}$ into a $d$-dimensional embedding $\varphi({\tt v}_*)\in\mathbb{R}^d$ so that any edge $\forall{\tt e}_*\in{\tt E}$ can be calculated therefrom. Specifically, we employ {\it evolutionary algorithm} (EA) to tune the EUD, which enjoys advantages over conventional gradient methods in both convergence speed and accuracy.

\begin{equation}
\label{eq:2}
    \arg\max\sum_{\tt t}\sum_{j=1}^n\log\text{Pr}({\tt r}_j^{\tt t}\in\bar{{\tt R}}_{\tt t}\mid\text{EUD})
\end{equation}

\noindent {\bf EA Setup} We use an array of real numbers ${\tt x}_{1:10}$ to code an individual in the population, where ${\tt x}_j\in[0,1]$ denotes the usefulness of $j$-th edge type. Given an EUD, {\it PageRank} \citep{Page1998The} runs on graph to infer the relative importance of each node for each target document, and a fitness function is applied to judge how well this EUD satisfies locating the ground truth references as Eq. \ref{eq:3}, in which if ${\tt r}_j^{\tt t}$ belongs to $\bar{{\tt R}}_{\tt t}$, then $\alpha({\tt r}_j^{\tt t},\bar{{\tt R}}_{\tt t})\in\mathbb{N}$ returns the ranking of ${\tt r}_j^{\tt t}$ within $\bar{{\tt R}}_{\tt t}$, and otherwise a big penalty coefficient to prevent irrelevant references to be recommended. Like most other optimizations, this procedure starts with a randomly generated population.

\begin{equation}
\label{eq:3}
    \max\dfrac{1}{\sum_{\tt t}\sum_{j=1}^n\left|j-\alpha({\tt r}_j^{\tt t},\bar{{\tt R}}_{\tt t})\right|}
\end{equation}

\noindent {\bf EA Operator} We choose the operator from {\it differential evolution} \citep{Das2011Differential} to generate offsprings for each individual. The basic idea is to utilize the difference between different individuals to disturb each trial object. First, three distinct individuals ${\tt x}_{1:10}^{r_1},{\tt x}_{1:10}^{r_2},{\tt x}_{1:10}^{r_3}$ are sampled randomly from current population to create a variant ${\tt x}_{1:10}^\text{var}$, as shown in Eq. \ref{eq:4}, where $f\in\mathbb{R}$ indicates the scaling factor. Next, ${\tt x}_{1:10}^\text{var}$ is crossed with a trial object ${\tt x}_{1:10}^\text{tri}$ to build a hybrid one ${\tt x}_{1:10}^\text{hyb}$ as Eq. \ref{eq:5}, in which $c\in[0,1]$ denotes the crossover factor and $u\in[0,1]$ represents an uniform random number. At last, the fitnesses of ${\tt x}_{1:10}^\text{tri}$ and ${\tt x}_{1:10}^\text{hyb}$ are compared, and the better one will be saved as the offspring into a new round of evolution. 

\begin{equation}
\label{eq:4}
    {\tt x}_j^\text{var}={\tt x}_j^{r_1}+f\times({\tt x}_j^{r_2}-{\tt x}_j^{r_3})
\end{equation}

\begin{equation}
\label{eq:5}
    {\tt x}_j^\text{hyb}=\begin{cases}
        {\tt x}_j^\text{var}, & \text{if}\quad u\leq c \\[2mm]
        {\tt x}_j^\text{tri}, & \text{otherwise}
    \end{cases}
\end{equation}

\subsection{Neural Extractive Summarization}

\begin{figure*}[htb]
\centering
    \includegraphics[width=0.8\textwidth]{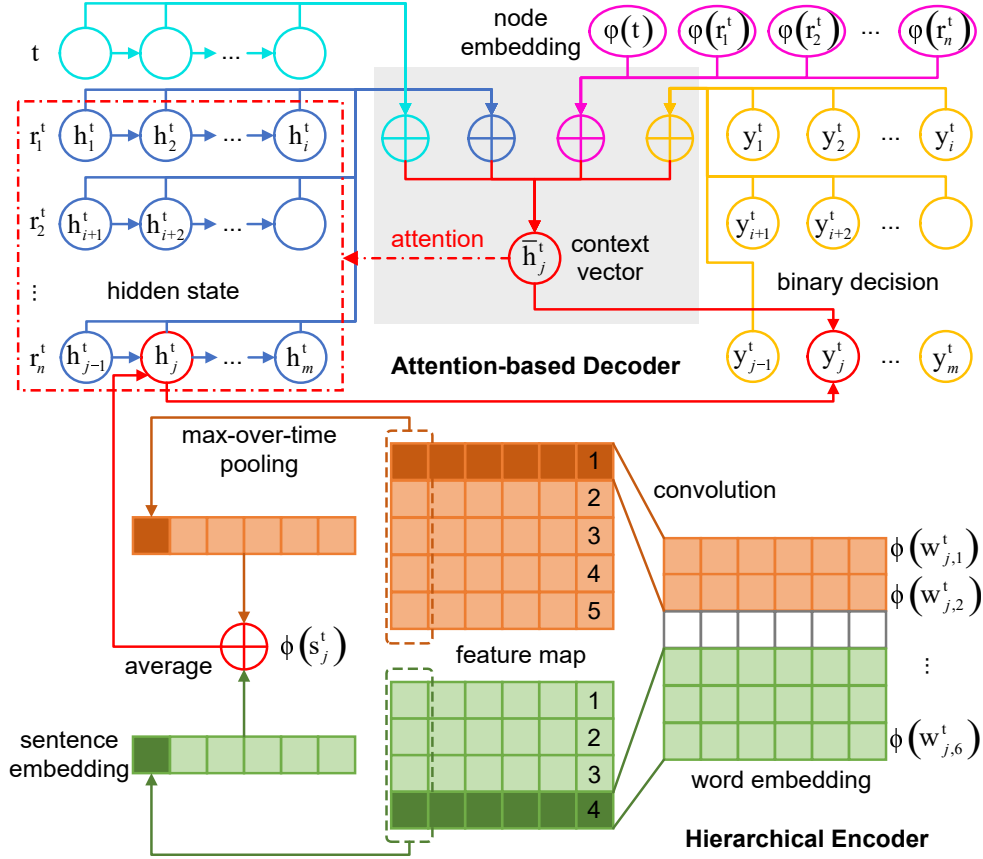}
\caption{Framework of our seq2seq summarizer.}
\label{fig:2}
\end{figure*}

As Figure \ref{fig:2} shows, we model our seq2seq summarizer with a hierarchical encoder and an attention-based decoder, as described below.

\noindent {\bf Hierarchical Encoder} Our encoder consists of two major layers, namely a {\it convolutional neural network} (CNN) and a {\it long-short-term memory} (LSTM)-based RNN. Specifically, the CNN deals with word-level texts to derive sentence-level meanings, which are then taken as inputs to the RNN for handling longer-range dependency within lager units like a paragraph and even a whole paper. This conforms to the nature of document that is composed from words, sentences and higher levels of abstraction \citep{Narayan2017Neural}.

Consider a sentence of $p$ words ${\tt s}_j^{\tt t}=\{{\tt w}_{j,1:p}^{\tt t}\}$, where each word ${\tt w}_{j,i}^{\tt t}$ can be represented by a $d$-dimensional embedding $\phi({\tt w}_{j,i}^{\tt t})\in\mathbb{R}^d$. Previous studies have illustrated the strength of CNN in presenting sentences, because of its capability to learn compressed expressions and address sentences with variable lengths \citep{Kim2014Convolutional}. First, a convolution kernel $k\in\mathbb{R}^{d\times q\times d}$ is applied to each possible window of $q$ words to construct a list of feature maps as: 

\begin{equation}
\label{eq:6}
    {\tt g}_{j,i}^{\tt t}=\tanh\big(k\times\phi({\tt w}_{j,i:i+q-1}^{\tt t})+b\big)
\end{equation}

\noindent where $b\in\mathbb{R}^d$ denotes the bias term. Next, {\it max-over-time pooling} \citep{Collobert2011Natural} is performed on all generated features to obtain the sentence embedding as:

\begin{equation}
\label{eq:7}
    \phi({\tt s}_j^{\tt t})=\mathop{\max}_{1\leq i\leq d}\big({\tt g}_{j,1:p-q+1}^{\tt t}[i,:]\big)
\end{equation}

\noindent where $[i,:]$ denotes the $i$-th row of matrix. Given a sequence of sentences ${\tt S}_{\tt t}=\{{\tt s}_{1:m}^{\tt t}\}$, we then take the RNN to yield an equal-length array of hidden states, in which LSTM has proved to alleviate the vanishing gradient problem when training long sequences \citep{Hochreiter1997Long}. Each hidden state can be viewed as a local representation with focusing on current and former sentences together, which is updated as: ${\tt h}_{j}^{\tt t}=\text{LSTM}\big(\phi({\tt s}_j^{\tt t}),{\tt h}_{j-1}^{\tt t}\big)\in\mathbb{R}^d$. 

In practice, we use multiple kernels with various widths to produce a group of embeddings for each sentence, and average them to capture the information inside different $n$-grams. As Figure \ref{fig:2} (bottom) shows, the sentence ${\tt s}_j^{\tt t}$ involves six words, and two kernels of widths two (orange) and three (green) abstract a set of five and four feature maps respectively. Meanwhile, since {\it rhetorical structure theory} \citep{Mann2009Rhetorical} points out that association must exist in any two parts of coherent texts, RNN is only applicable to manage the sentence relation within a single document, because we cannot expect the dependency between two sections from different references.

\noindent {\bf Attention-based Decoder} Our decoder labels each sentence ${\tt s}_j^{\tt t}$ as 0/1 sequentially, according to whether it is salient or novel enough, plus if relevant to the target document ${\tt t}$ or not. As shown in Figure \ref{fig:2} (top), the binary decision ${\tt y}_j^{\tt t}$ is made by both the hidden state ${\tt h}_j^{\tt t}$ and the context vector $\bar{{\tt h}}_j^{\tt t}$ from an attention mechanism (grey background). In particular, this attention (red dash line) is acted as an intermediate stage to determine which sentences to highlight so as to provide the contextual information for current decision \citep{Bahdanau2014Neural}. Given ${\tt H}_{\tt t}=\{{\tt h}_{1:m}^{\tt t}\}$, this decoder returns the probability of ${\tt y}_j^{\tt t}=1$ as below:

\begin{equation}
\label{eq:8}   
    \text{Pr}({\tt y}_j^{\tt t}=1\mid{\tt R}_{\tt t};{\tt S}_{\tt t};\theta)=\text{sigmoid}\big(\delta({\tt h}_j^{\tt t},\bar{{\tt h}}_j^{\tt t})\big)
\end{equation}

\begin{equation}
\label{eq:9}  
    \bar{{\tt h}}_j^{\tt t}=\sum_{i=1}^m{\tt a}_{j,i}{\tt h}_i^{\tt t}
\end{equation}

\noindent where $\delta({\tt h}_j^{\tt t},\bar{{\tt h}}_j^{\tt t})\in\mathbb{R}$ denotes a fully connected layer with as input the concatenation of ${\tt h}_j^{\tt t}$ and $\bar{{\tt h}}_j^{\tt t}$, and ${\tt a}_{j,i}\in[0,1]$ is the attention weight indicating how much the supporting sentence ${\tt s}_i^{\tt t}$ contributes to extracting the candidate one ${\tt s}_j^{\tt t}$.

Apart from saliency and novelty two traditional attention factors \citep{Chen2016Distraction,Tan2017Abstractive}, we focus on the contextual relevance within both textual and graphic contexts to distinguish the relationship from near to far, as shown in Eq. \ref{eq:10} and Eq. \ref{eq:11}. To be specific: 1) ${\tt h}_j^{{\tt t}\mathrm{T}}W_{\tt s}{\tt h}_i^{\tt t}$ represents the saliency of ${\tt s}_i^{\tt t}$ to ${\tt s}_j^{\tt t}$; 2) $-{\tt d}_j^{{\tt t}\mathrm{T}}W_{\tt n}{\tt h}_i^{\tt t}$ indicates the novelty of ${\tt s}_i^{\tt t}$ to the dynamic output ${\tt d}_j^{{\tt t}}$; 3) $\phi({\tt t})^\mathrm{T}W_{\tt t}{\tt h}_i^{\tt t}$ denotes the relevance of ${\tt s}_i^{\tt t}$ to ${\tt t}$ from the textual context; 4) $\varphi({\tt t})^\mathrm{T}W_{\tt g}\varphi({\tt h}_i^{\tt t})$ refers to the relevance from the graphic context. More concretely, $W_*\in\mathbb{R}^d$ characterizes the learnable matrix, $\phi({\tt t})$ returns the average of hidden states from ${\tt t}$, $\varphi({\tt t})$ and $\varphi({\tt h}_i^{\tt t})$ return the node embeddings of both ${\tt t}$ and the source document that ${\tt h}_i^{\tt t}$ belongs to respectively. Note that $\phi(\cdot)$ and $\varphi(\cdot)$ represent two distinct embedding spaces, where the former reflects the lexical collocations of corpus, and the latter embodies the connectivity patterns of associated graph.

\begin{equation}
\label{eq:10}
    \begin{split}
        {\tt a}_{j,i}=\quad\quad{\tt h}_j^{{\tt t}\mathrm{T}}W_{\tt s}{\tt h}_i^{\tt t}&\quad\text{\# saliency}\\[2mm]
        -{\tt d}_j^{{\tt t}\mathrm{T}}W_{\tt n}{\tt h}_i^{\tt t}&\quad\text{\# novelty}\\[2mm]
        +\phi({\tt t})^\mathrm{T}W_{\tt t}{\tt h}_i^{\tt t}&\quad\text{\# relevance}_1\\[2mm]
        +\varphi({\tt t})^\mathrm{T}W_{\tt g}\varphi({\tt h}_i^{\tt t})&\quad\text{\# relevance}_2
    \end{split}
\end{equation}

\begin{equation}
\label{eq:11}   
    {\tt d}_j^{\tt t}=\sum_{i=1}^{j-1}\text{Pr}({\tt y}_j^{\tt t}=1\mid{\tt R}_{\tt t};{\tt S}_{\tt t};\theta)\times{\tt h}_i^{\tt t}
\end{equation}

The basic idea behind our attention mechanism is as follows: if a supporting sentence more resembles a candidate one, or overlaps less with the dynamic output, or is more relevant to the target document, then it can provide more contextual information to facilitate current decision on being extracted or not, thereby taking a higher weight in the generated context vector. This innovative attention will guide our goal related work section to maximize the representativeness of selected sentences (saliency \& novelty), while minimizing the semantic distance to the target document (relevance). This is consistent with the way that scholars consume a reference collection, with the minmax objective in their minds.
\section{Experiment}

\subsection{Experimental Setup}

This section presents the experimental setup for assessing our approach, including 1) dataset used for training and testing, 2) implementation details, 3) contrast methods and evaluation metrics.

\noindent{\bf Dataset} We conduct experiments on a dataset\footnote{To help readers reproduce the experiment outcome, we share part of the experiment data while the copyrighted information is removed. \url{https://github.com/kuadmu/2018EMNLP}} created from the ACM digital library, where metadata and full texts are derived from PDF files. To be detailed, this dataset includes 371,891 papers, 779,810 authors, 9,204 keywords and 807 venues in total. Note that we ignore the keyword with frequency below a certain threshold, and adopt {\it greedy matching} of \citet{Guo2013Scientific} to generate pseudo keywords for papers lacking topic descriptions. For each target document, the references are traversed by the descending order of the cited number in related work section (primary) and in full paper (secondary) successively. We first apply a series of pre-processings such as lowercasing and stemming to standardize candidate sentences, then remove those which are too short/long ($<7$ or $>80$ words). On this basis, a total of 8,080 papers are selected to evaluate our approach, each containing more than 15 references found in the dataset and a related work section of at least 500 words. But as for the heterogeneous bibliography graph, all source data have to be imported to ensure the structural integrity of communities. Besides, this graph should be constructed year-by-year to preclude the effect of later publications on earlier ones.

\noindent{\bf Implementation} We use {\it Tensorflow} for implementation, where both the dimensions of embedding and hidden state are equally 128. For the CNN, {\it word2vec} \citep{Mikolov2013Efficient} is utilized to initialize the word embeddings, which can be further tuned during the training phase. Meanwhile, we follow the work of \citet{Kim2014Convolutional} to apply a list of kernels with widths $\{3,4,5\}$. As for the RNN, each LSTM module is set to one single layer, and all input documents are padded to the same length, along with a mark to indicate the real number of sentences. Based on these settings, we train our summarizer using {\it Adam} with the default in \citet{Kingma2014Adam}, and perform {\it mini-batch cross-entropy} training with a batch of one target document for 20 epochs.

To create training data for our summarizer, each reference needs to be annotated with the ground truth in advance, i.e., candidate sentences are tagged with 0/1 for indicating summary-worthy or not. Specifically, we follow a heuristic practice of \citet{Cao2016AttSum} and \citet{Nallapati2016SummaRuNNer} to compute ROUGE-2 score \citep{Lin2003Automatic} for each sentence, in terms of the native related work sections (gold standards). Next, those sentences with high scores are chosen as the positive samples, and the rest as the negative ones, such that the total score of selected sentences is maximized with respect to the gold standard. As for testing, we relax the number of sentences to be selected, and focus on the classification probability from Eq. \ref{eq:8}. In this study, {\it cross validation} is applied to split the dataset into ten parts equally at random, in which nine are used for training and the other one for testing.

\noindent{\bf Evaluation} We adopt the widely used toolkit ROUGE \citep{Lin2003Automatic} to evaluate the summarization performance automatically. In particular, we report ROUGE-1 and ROUGE-2 (unigram and bigram overlapping) as a way to assess the informativeness, and ROUGE-L (the longest common subsequence) as a means to assess the fluency, in terms of fixed bytes of gold standards.

To validate the proposed attention mechanism, we compare our approach (denoted as $\text{P.}_\text{S+N+Rteg+EUD}$) against six variants, including: 1) $\text{P.}_\text{void}$: a plain seq2seq summarizer without attentions; 2) $\text{P.}_\text{S}$: use the saliency as an only attention factor; 3) $\text{P.}_\text{S+N}$: leverage both the saliency and novelty; 4) $\text{P.}_\text{S+N+Rt}$: incorporate the relevance from the textual context; 5) $\text{P.}_\text{S+N+Rtog}$: gain the relevance from the graphic context of a homogeneous citation graph; 6) $\text{P.}_\text{S+N+Rteg}$: utilize the heterogeneous bibliography graph, but with each edge type the same usefulness.

In addition, we also select six representative summarization methods as a benchmark group. The first one is the general seq2seq summarizer by \citet{Cheng2016Neural}, denoted as PointerNet, which employs an attention mechanism to extract sentences directly after reading them. Following are five classical generic solutions, including: 1) Luhn \citep{Luhn1958The}: a heuristic summarization based on word frequency and distribution; 2) MMR \citep{Carbonell1998MMR}: a diversity-based re-ranking to produce summaries; 3) LexRank \citep{Erkan2004LexRank}: a graph-based summary technique inspired by {\it PageRank} and {\it HITS}; 4) SumBasic \citep{Nenkova2005The}: a frequency-based summarizer with duplication removal; 5) NltkSum \citep{Acanfora2014Natural}: a {\it natural language tookit} (NLTK)-based implementation for summarization.

For clarity, Luhn, LexRank and SumBasic are analogous to the work of \citet{Hu2014Automatic} which extracts sentences scoring the highest in significance, and they are also contrasted in the latest studies on neural summarizers \citep{Chen2016Distraction,Tan2017Abstractive}. Meanwhile, MMR often serves as a part/post-processing of existing techniques to avoid the redundancy \citep{Cohan2017Scientific}, and we introduce NltkSum to investigate the impact of grammatical/semantic analysis to the automatic related work summarization. Note that former studies specially for this task require extensive human involvements (see Table \ref{table:1}), thus we cannot apply them to such a large dataset of this study.

\subsection{Results and Discussion}

\begin{table*}[t!]
    \begin{center}
        \begin{tabular}{|l|lll|}
        \hline Methods & ROUGE-1 & ROUGE-2 & ROUGE-L \\ \hline
        $\text{P.}_\text{void}$ & 26.85* & 6.38* & 14.22* \\
        $\text{P.}_\text{S}$ & 26.98* & 6.48* & 14.36* \\
        $\text{P.}_\text{S+N}$ & 27.29* & 6.65* & 14.43* \\
        $\text{P.}_\text{S+N+Rt}$ & 27.63* & 6.72* & 14.46* \\
        $\text{P.}_\text{S+N+Rtog}$ & 27.82* & 7.00* & 14.55* \\
        $\text{P.}_\text{S+N+Rteg}$ & 28.56* & 7.40 & 14.70* \\
        $\text{P.}_\text{S+N+Rteg+EUD}$ & \bf 29.18 & \bf 7.63 & \bf 14.89 \\
        \hline 
        Luhn & 25.76* & 5.08* & 13.50* \\
        MMR & 25.55* & 5.14* & 13.99* \\
        LexRank & 25.07* & 5.12* & 13.95* \\
        SumBasic & 28.01* & 5.44* & 13.93* \\
        NltkSum & 28.07* & 6.36* & 14.87 \\
        PointerNet & 27.06* & 6.53* & 14.41* \\
        \hline
        \end{tabular}
        \begin{tablenotes}
            * indicates {\it Wilcoxon signed-rank test} $p<0.01$, compared with $\text{P.}_\text{S+N+Rteg+EUD}$        
        \end{tablenotes}
    \end{center}
\caption{\label{table:2} Rouge evaluation (\%) on 8,080 papers from ACM digital library.}
\end{table*}

Table \ref{table:2} reports the evaluation comparison over ROUGE metrics. From the top half, all scores appear a gradual upward trend with incorporation of saliency, novelty, relevance (from both textual and graphic contexts) and EUD into consideration one after another, which demonstrates the validity of our attention mechanism for summarizing related work sections. To be specific, we further reach the following conclusions: 

1) $\text{P.}_\text{void}$ vs. $\text{P.}_\text{S}$ vs. $\text{P.}_\text{S+N}$: Both saliency and novelty are two effective factors to locate the required content for summaries, which is consistent with prior studies. 

2) $\text{P.}_\text{S+N}$ vs. $\text{P.}_\text{S+N+Rt}$: Contextual relevance does contribute to address the alignment between a related work section and its source documents.

3) $\text{P.}_\text{S+N+Rt}$ vs. $\text{P.}_\text{S+N+Rtog}$: Textual context alone cannot provide entire evidence to characterize the relationship among scientific publications exactly.

4) $\text{P.}_\text{S+N+Rtog}$ vs. $\text{P.}_\text{S+N+Rteg}$: Heterogeneous bibliography graph involves richer contextual information than a homogeneous citation graph.

5) $\text{P.}_\text{S+N+Rteg}$ vs. $\text{P.}_\text{S+N+Rteg+EUD}$: EUD plays an indispensable role in organizing accurate contextual relevance on a heterogeneous graph.

\begin{figure}[htb]
\centering
    \includegraphics[width=0.5\textwidth]{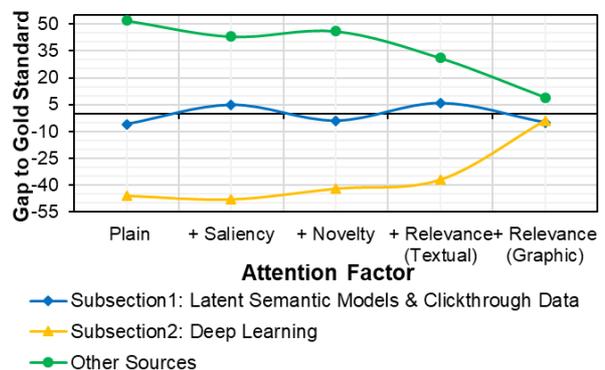}
\caption{Number of extracted words on each reference cluster under different attention factors.}
\label{fig:3}
\end{figure} 

Continuing the ``DSSM", Figure \ref{fig:3} visualizes the number of extracted words on each reference cluster\footnote{We pack the references cited in the same subsection of the related work section as one reference cluster.} under different attention factors. It can be seen that only after adding the relevance especially that from the graphic context into attentions, our summarizer can correctly sample the content from ``Deep Learning" (yellow line), and eliminate that originated from ``Other Sources" by a big margin (green line). As this example falls into the methodology transferring, a host of its involved word collocations are not idiomatic combinations yet, such as ``Deep Neural Network" co-occurs with ``Clickthrough Data" that is more frequently related to ``Latent Semantic Analysis" at that time, which results in a somewhat biased textual context. By contrast, the graphic context will suffer less from this bias because it characterizes the connectivity patterns (real-time setup) instead of $n$-gram statistics, thus offering a more robust measure for the contextual relevance.

The bottom half of Table \ref{table:2} illustrates the superiority of our approach over six representative summarization methods. Above all, Luhn, LexRank and MMR three summarizers that simply exploit shallow text features (word frequency and associated sentence similarity) to measure either significance or redundancy fall far behind the plain variant $\text{P.}_\text{void}$, which partly reflects the strength of seq2seq paradigm in summarizing a related work section. Second, with combination of significance and redundancy, SumBasic achieves a drastic increase on ROUGE-1 and a mild raise on ROUGE-2 respectively, but it still cannot improve ROUGE-L marginally. This is because simple text statistics cannot present deeper levels of natural language understanding to catch larger-grained units of co-occurrence. Third, NltkSum benefits from a NLTK library so as to access grammatical/semantic supports, thereby having the best informativeness (ROUGE-1 and ROUGE-2) among the five generic baselines, and meanwhile a comparable fluency (ROUGE-L) with our approach. Finally, as a deep learning solution, although PointerNet takes both hidden states and previously labeled sentences into account, at each decoding step it focuses on only current and just one previous sentences, lacking a comprehensive consideration on saliency, novelty and more importantly the contextual relevance ($<\text{P.}_\text{S+N}$).

To better verify the summarization performance, we also conduct a human evaluation on 35 papers containing more than 30 references in the dataset. We assign a number of raters to compare each generated related work section against the gold standard, and judge by three independent aspects as: 1) How compliant is the related work section to the target document? 2) How intuitive is the related work section for readers to grasp the key content? 3) How useful is the related work section for researchers to prepare their final literature reviews? Note that we do not allow any ties during the comparison, and each property is assessed with a 5-point scale of 1 (worst) to 5 (best).

\begin{table*}[t!]
    \begin{center}
        \begin{tabular}{|l|lllllllc|}
        \hline Methods & 1st & 2nd & 3rd & 4th & 5th & 6th & 7th & Mean Ranking \\ \hline
        Luhn & 0.04 & 0.07 & 0.09 & 0.13 & 0.17 & 0.23 & 0.29 & 5.26 \\
        MMR & 0.05 & 0.07 & 0.11 & 0.16 & 0.19 & 0.22 & 0.20 & 4.82 \\
        LexRank & 0.06 & 0.09 & 0.11 & 0.14 & 0.17 & 0.19 & 0.27 & 4.93 \\
        SumBasic & 0.09 & 0.13 & 0.18 & 0.18 & 0.18 & 0.15 & 0.10 & 4.10 \\
        NltkSum & 0.21 & 0.21 & 0.20 & 0.15 & 0.10 & 0.07 & 0.04 & 3.00 \\
        PointerNet & 0.14 & 0.20 & 0.18 & 0.15 & 0.13 & 0.11 & 0.08 & 3.54 \\
        $\text{P.}_\text{S+N+Rteg+EUD}$ & 0.40 & 0.22 & 0.14 & 0.09 & 0.06 & 0.04 & 0.02 & 2.34 \\
        \hline
        \end{tabular}
    \end{center}
\caption{\label{table:3} Human evaluation (proportion) on 35 papers with more than 30 references in the dataset.}
\end{table*} 

Table \ref{table:3} displays how often raters rank each summarizer as the 1st, 2nd and so on, in terms of best-to-worst. Specifically, our approach comes the 1st on 40\% of the time, which is followed by NltkSum that is considered the best on 21\% of the time (almost half of ours), and PointerNet with quite equal proportions on each ranking. Furthermore, the other four summarizers account for obviously lower ratings in general. To attain the statistical significance, {\it one-way analysis of variance} (ANOVA) is performed on the obtained ratings, and the results show that our approach is better than all six contrast methods significantly ($p<0.01$), which means that the conclusion drawn by Table \ref{table:2} is sustained.
\section{Conclusion}

In this paper, we highlight the contextual relevance for the automatic related work summarization, and analyze the graphic context to characterize the relationship among scientific publications accurately. We develop a neural data-driven summarizer by leveraging the seq2seq paradigm, where a joint context-driven attention mechanism is proposed to measure the contextual relevance within full texts and a heterogeneous bibliography graph simultaneously. Extensive experiments demonstrate the validity of the proposed attention mechanism, and the superiority of our approach over six representative summarization baselines.

In future work, an appealing direction is to organize the selected sentences in a logical fashion, e.g., by leveraging a topic hierarchy tree to determine the arrangement of the related work section \citep{Cong2010Towards}. We also would like to take the citation sentences of each reference into consideration, which is another concise and universal data source for scientific summarization \citep{Chen2016Summarization,Cohan2017Scientific}. At the end of this paper, we believe that extractive methods are by no means the final solutions for literature review generation due to plagiarism concerns, and we are going to put forward a fully abstractive version in further studies.

\section*{Acknowledgement}
We would like to thank the anonymous reviewers for their valuable comments. This work is partially supported by the National Science Foundation of China under grant No. 71271034.

\bibliography{emnlp2018}
\bibliographystyle{acl_natbib}

\end{document}